# On the Study of Biometric Spoofing Detection using Deep Learning

Kumar Kartikey and Nikos Komninos

School of Science and Technology
City St George's University of London,
London EC10V, UK

*Abstract*— **Biometric systems are increasingly deployed in security applications; however, they remain vulnerable to spoofing attacks, wherein attackers exploit counterfeit biometric data to gain unauthorized access. This research evaluates the effectiveness of state-of-the-art machine learning models—MobileNetV2, DenseNet-121, Inception-v3, and Spoof Trace Disentanglement (STD)—in detecting spoofing attacks within facial recognition systems. Using the CelebA-Spoof dataset, the study evaluates model performance using metrics such as accuracy, precision, recall, and F1 Score. Cross-dataset validation is performed on the MSU-MFSD dataset to assess generalizability. The results reveal MobileNetV2 as the most efficient model, achieving 92% accuracy while balancing computational efficiency, making it suitable for real-world applications. Inception-v3 shows moderate robustness, while DenseNet-121 and STD experience challenges with generalization. The findings emphasize the need for advancements in domain adaptation and hybrid architectures to enhance biometric security systems.**

## I. Introduction

Biometric systems have become a cornerstone of modern security applications, offering greater convenience and security than traditional password-based systems. These systems rely on unique physiological or behavioral traits, such as facial features, fingerprints, and iris patterns, to verify identity. Among these, *facial recognition* has gained significant popularity due to its non-intrusive nature and ease of use (Zhang et al., 2020). However, the widespread adoption of biometric systems has also made them targets for *spoofing attacks*, in which attackers use counterfeit biometric data to deceive the system.

Spoofing attacks in facial recognition systems can involve printed photographs, replayed videos, or sophisticated 3D masks. These attacks exploit the inability of some systems to differentiate between live and spoofed inputs, compromising system integrity. Successful spoofing attacks have severe implications, ranging from unauthorized access to sensitive information to financial losses and compromised security in critical infrastructure (Yang et al., 2023). As spoofing techniques evolve, there is an urgent need for robust and adaptable detection mechanisms.

The motivation for this research stems from the growing sophistication of spoofing techniques and the limitations of existing detection methods. Traditional anti-spoofing approaches, such as hardware-based solutions and feature-engineered software methods, often struggle to adapt to novel attacks and require specialized hardware, limiting their scalability. Machine learning, particularly deep learning, has emerged as a promising solution due to its ability to learn complex patterns and adapt to diverse spoofing scenarios. However, challenges such as domain adaptation, adversarial robustness, and computational efficiency remain unresolved.

This research addresses these challenges by evaluating the effectiveness of advanced machine learning models in detecting biometric spoofing attacks. The study focuses on facial recognition systems and leverages the *CelebA-Spoof* dataset, a large-scale, richly annotated resource, to train and evaluate models. Cross-dataset validation is conducted using the *MSU-MFSD* dataset to assess generalizability. The findings contribute to the development of secure and reliable biometric systems capable of withstanding evolving spoofing threats.

## II. Related Studies

Research in biometric spoof detection has significantly evolved over the past decade, transitioning from handcrafted feature-based techniques to deep learning-driven methods. Various machine learning models have been employed to detect presentation attacks, with Convolutional Neural Networks (CNNs) demonstrating superior performance in feature extraction. The effectiveness of biometric spoo- detection methods has been evaluated using benchmark datasets such as *MSU-MFSD, OULU-NPU, CASIA-SURF, and CelebA-Spoof,* which provide diverse attack scenarios for robust model training and evaluation.

*A. Traditional Approaches*

Previous methods for biometric spoof detection focused on handcrafted feature extraction techniques. For example, Local Binary Patterns (LBP) were applied for texture analysis, and LBP effectively distinguished between live and spoofed images in controlled conditions (Määttä et al., 2011). However, LBP faced challenges due to variations in lighting and pose, as well as complex spoofing techniques. *Moreover,* Histogram of Oriented Gradients (HOG) successfully captured edge and shape information but was ineffective in detecting high-quality presentation attacks, such as *3D masks* and replay attacks (Patel et al., 2016). Motion analysis and optical flow techniques, such as eye-blink detection (Pan et al., 2007) and *lip* movement analysis (Li et al., 2016), were investigated for liveness detection. While effective at identifying static face spoofing, these methods struggled against high-fidelity video-based attacks.

While traditional approaches were computationally efficient, they were not adaptable to real-world spoofing conditions, limiting their applicability in dynamic authentication systems.

*B. Deep Learning Approaches*

The emergence of deep learning, especially CNNs, has revolutionized biometric spoof detection by facilitating automated feature extraction from raw image data. Numerous studies have demonstrated the advantages of CNN-based methods over traditional feature-based techniques.

*Inception-v3 (Szegedy et al., 2016):* Demonstrated high accuracy in facial spoof detection by capturing multi-scale spatial patterns. However, its computational requirements rendered it unsuitable for real-time applications.

*DenseNet-121 (Huang et al., 2017):* Leveraged dense connections between layers to enhance feature reuse and gradient flow, leading to improved performance on large-scale datasets. However, it continued to be computationally intensive.

*MobileNetV2 (Sandler et al., 2018):* Designed for efficiency, MobileNetV2 achieved competitive accuracy while significantly reducing computational costs, making it well-suited for mobile and embedded systems.

Several dataset-driven studies have further advanced the field. For example, *Zhang et al. (2020)* introduced the *CelebA-Spoof dataset,* which includes a variety of attack types such as *2D print attacks, video replay attacks, and 3D mask attacks*, significantly enhancing model robustness.

*Liu et al. (2024)* examined issues of cross-dataset generalization, highlighted the *domain shift problem* in biometric spoof detection, and advocated for domain adaptation techniques. *Yang et al. (2023)* explored *adversarial robustness,* demonstrating that CNNs are vulnerable to adversarial attacks in which imperceptible perturbations can deceive even high-performing models. *George et al. (2019)* proposed a multi*modal approach* that combines *RGB, depth, and infrared (IR) data,* resulting in a more robust defence against spoofing attacks.

**Table 1: Pros and Cons**

| ***Approach*** | ***Pros*** | ***Cons*** |
|---|---|---|
| ***Handcrafted Feature Methods*** | Computationally efficient, interpretable, works well under controlled conditions | Poor generalization, ineffective against sophisticated attacks |
| ***CNN-Based Approaches*** | High accuracy, automatic feature extraction, robust against common spoofing attacks | Computationally expensive, requires large datasets, vulnerable to adversarial attacks |
| ***Lightweight Models (MobileNetV2, EfficientNetB0)*** | Efficient, suitable for real-time applications | Slightly lower accuracy compared to deeper models |
| ***Domain Adaptation Methods*** | Improved cross-dataset generalization | Requires unsupervised adaptation and additional computational resources |
| ***Multi-Modal Approaches (RGB + Depth + IR)*** | More robust against spoofing techniques | Requires specialized hardware (depth sensors, IR cameras) |
| ***Adversarial Training Techniques*** | Enhances robustness against adversarial examples | Computationally intensive, requires specialized training data |

### C. Limitations and Conclusions from Existing Studies

Despite significant progress, several limitations persist in biometric spoof detection.

#### I. Cross-Dataset Generalization

Many models perform well on the datasets they are trained on but have *difficulty generalizing* when evaluated on unseen datasets. The domain gap between controlled datasets and real-world conditions reduces detection accuracy.

#### II. Vulnerability to Adversarial Attacks

CNN-based models are vulnerable *to adversarial perturbations* that can mislead even the most accurate classifiers. Research by *Xu et al. (2022)* suggests the need for *adversarial training* and *defence mechanisms* to enhance robustness.

#### III. Computational Complexity vs. Efficiency Trade-Off

While deeper models like *DenseNet-121* offer higher accuracy, they require substantial computational *power,* making them impractical for real-time applications. Lightweight models (e.g., MobileNetV2, EfficientNet) provide an efficient alternative but at the cost of slightly reduced accuracy.

Most datasets focus on standard spoofing techniques, but *new attack vectors* (e.g., deepfake-based biometric attacks) remain underrepresented.

## III. Exploration of Deep Learning Models

This research presents an in-depth evaluation of four state-of-the-art deep learning models for biometric spoof detection: Inception-v3, DenseNet-121, MobileNetV2, and Spoof Trace Disentanglement (STD). Each model has been chosen for its distinct capability to capture variations in facial texture, identify spoof artifacts, and balance computational efficiency with accuracy. The primary objective of this research is to develop an approach that achieves high performance in biometric spoof detection while ensuring practical applicability in real-world scenarios, particularly in mobile and resource-constrained environments.

### A. Design Approach

The proposed approach involves a structured machine learning pipeline that encompasses data preprocessing, model selection, training, validation, and evaluation. A key focus is to ensure robustness against various spoofing techniques while maintaining computational efficiency. The workflow includes the following essential stages:

1. *Data Preprocessing:* Raw images are standardized by resizing, normalizing, and applying augmentation techniques to enhance model robustness.
2. *Model Selection and Implementation:* ***Various*** CNN architectures are evaluated based on their effectiveness in feature extraction and classification.
3. *Training and Optimization:* Each model is trained with a structured loss function and adaptive optimization algorithms.
4. *Evaluation and Benchmarking:* Performance metrics, including accuracy, precision, recall, F1-score, and True Positive Rate (TPR) at designated False Positive Rate (FPR) thresholds, are analyzed.
5. *Cross-Dataset Validation:* Generalization ability is assessed by evaluating models on unseen datasets, such as the MSU-MFSD dataset.

This methodology guarantees that the proposed approach is thorough, balancing accuracy and computational demands to ease deployment in real-world security systems.

### B. Architectural Approach

The foundation of this research lies in convolutional neural networks (CNNs), which have demonstrated exceptional performance in facial analysis and biometric recognition. The framework comprises multiple interconnected layers, each playing a vital role in the spoof detection process:

- *Input Layer:* The model receives preprocessed facial images from the CelebA-Spoof dataset, ensuring uniformity in dimensions and intensity levels.

- *Feature Extraction Layers:* CNN-based architectures extract spatial features that distinguish real images from spoofed ones. Convolutional layers apply multiple filters to capture textural inconsistencies, while pooling layers reduce dimensionality.
- *Classification Layer:* Fully connected layers evaluate extracted features and assign a classification label (real/spoof). The final SoftMax layer assigns probability to each class.
- *Decision Thresholding:* To reduce false positives, a predefined threshold is established to maintain high sensitivity to spoof attacks while minimizing misclassification.

### *C. Model Selection*

Four deep learning architectures were selected based on their effectiveness in feature extraction and their computational trade-offs:

#### *1. Inception-v3:*

A widely used CNN architecture known for its inception modules, which enable multi-scale feature extraction. Effective in identifying subtle textural anomalies in facial images. It requires substantial computational resources, which makes it less suitable for real-time mobile deployment.

#### *2. DenseNet-121:*

It utilizes densely connected layers to improve feature reuse and gradient flow. Effective at extracting detailed facial features essential for identifying advanced spoofing techniques. Although robust, it has a large memory footprint, which can limit its use in low-power applications.

#### *3. MobileNetV2:*

A lightweight CNN designed for computational efficiency uses depth-wise separable convolutions to reduce processing time while maintaining competitive accuracy. It is ideal for mobile and edge-device applications, though slightly less accurate in highly complex spoofing scenarios.

#### *4. Spoof Trace Disentanglement (STD):*

A specialized model developed to isolate and analyse spoof-specific artefacts. It explicitly models spoof traces, providing improved interpretability in classification. However, it struggles with generalization across diverse datasets, particularly those with varied environmental conditions.

### *D. Dataset Preparation*

The **CelebA-Spoof dataset** serves as the primary dataset for training and evaluation. It consists of 625,537 images labeled with 43 attributes, covering a broad range of spoofing methods, including printed photos, 3D masks, and digital screen attacks. The dataset offers diversity in ethnicity, gender, lighting conditions, and camera perspectives, making it an ideal benchmark for biometric spoof detection.

To evaluate cross-dataset generalization, the **MSU-MFSD dataset** is incorporated for validation. This dataset primarily includes mobile-specific spoofing attacks, such as replay attacks via smartphones and tablets. Testing on an unseen dataset, the robustness of each model in handling real-world variability is assessed.

*Data Preprocessing Steps:*

- *Image Resizing:* Standardized to 224x224 pixels to ensure uniform input across all models.
- *Normalization:* Pixel values are scaled between [0, 1] to ensure stable convergence during training.
- *Data Augmentation:* Techniques like random rotation, brightness adjustment, and flipping are employed to enhance generalization and avoid overfitting.

### *E. Training Strategy*

To ensure optimal learning, the training strategy incorporates advanced techniques for model optimization and regularization. For example, *c*ross-entropy loss is employed to minimize classification errors. Adam optimizer with an adaptive learning rate is used to enhance convergence speed and stability. *D*ropout layers (set at 0.5) and batch normalization are implemented to reduce overfitting and improve generalization.
Moreover, 32 images per batch were selected to balance memory efficiency and training stability, and an initial learning rate of 0.001, with a decay factor of 0.9 every 10 epochs, was used to fine-tune weight updates. Finally, training was monitored to prevent unnecessary computation once the validation loss stabilized.

The dataset was divided into 70% for training, 20% for validation, and 10% for testing, using the **CelebA-Spoof dataset**. The evaluation phase assesses standard classification metrics, including the accuracy, which measures the percentage of correctly classified images*;* precision, which evaluates the proportion of true positives among all predicted positives, ensuring the model minimizes false alarms*;* recall, which assesses the model's ability to detect all spoof images correctly; F1-Score, which provides a harmonic mean of precision and recall balancing detection sensitivity and specificity; TPR@FPR, which measures the true positive rate at different false positive rate thresholds (1^e-3, 5^e-3, 1^e-4), providing insight into model performance under stringent security requirements.

Based on these metrics, the performance of all models is compared, and the results are visualized to facilitate comparative analysis.

### *G. Cross-Dataset Validation*

A critical aspect of the evaluation process is testing model generalization beyond the dataset on which it was trained. To achieve this, the trained models are evaluated on the **MSU-MFSD dataset**,whichg includes attack scenarios not encountered during training. The decline in performance across datasets underscores the need for domain adaptation techniques to enhance real-world applicability.

## IV. EXPERIMENTAL RESULTS

### *A. Testbed and Datasets*

The experimental evaluation of biometric spoof detection models was conducted using a combination of well-established datasets to ensure a thorough analysis. The primary training dataset was **CelebA-Spoof**, while cross-dataset validation was performed on the **MSU-MFSD** dataset. These datasets were chosen to encompass a wide array of spoofing techniques and authentic variations in biometric attacks.

The *CelebA-Spoof* dataset comprises 625,537 images from 10,177 subjects. It incorporates diverse spoofing techniques, including printed photo attacks, *3D masks*, and *replayed video* attacks. Due to its scale and variability, this extensive dataset is particularly suitable for training deep learning models.

The *MSU-MFSD* dataset, though smaller, serves as a cross-validation benchmark to assess models' generalization ability. It primarily includes *mobile-based attacks*, such as *printed photographs* and *video replays*, which makes it useful for evaluating real-world robustness.

These datasets enable us to evaluate not only the models' effectiveness in identifying spoofed biometric data but also their ability to generalise across diverse sources and attack methods.

### *B. Training Configuration*

The models were trained on the *Hyperion HPC cluster,* a high-performance computing environment that offered advanced hardware and software support for deep learning computations. The configurations were as follows: *Hardware: NVIDIA A100 GPUs* (40GB memory) for deep learning acceleration; *Intel Xeon Gold 6248R CPUs*( 48 cores per node) to handle large-scale parallel processing; *64GB* of allocated RAM per job, ensuring efficient execution of deep learning models.

Software: Python 3.9.7 was the primary programming language. PyTorch 2.5.1 and TensorFlow were used for deep learning models, along with essential libraries such as NumPy, Torchvision, and Scikit-learn for data preprocessing and performance evaluation.

These configurations ensured that training was conducted efficiently and that models utilised state-of-the-art hardware capabilities.

#### *1. Baseline Performance*

The baseline model, *ResNet-18*, achieved an accuracy of *89.5%* on the CelebA-Spoof dataset, with TPR@FPR values of *0.9864, 0.9730*, and *0.8912* at thresholds of 1e-3, 5e-3, and 1e-4, respectively. These results provide a benchmark for comparing advanced models.

#### *2. Model Comparison*

The assessment of biometric spoof detection models was conducted using key performance metrics, including accuracy, precision, recall, F1-score, and the True Positive Rate (TPR), across various False Positive Rates (FPR). Each model was assessed on its ability to distinguish genuine from spoofed biometric samples, while considering its computational efficiency, real-world applicability, and generalisation across different datasets.

*MobileNetV2* emerged as the most efficient and accurate model for biometric spoof detection, achieving an impressive accuracy of 92%, with a precision *96%* and a recall *93%.* The lightweight architecture makes it particularly well-suited for real-time applications, especially on mobile and embedded systems. The model's ability to process images quickly without compromising accuracy makes is suitable for large-scale authentication systems. However, despite its strong performance on the CelebA-Spoof dataset, MobileNetV2 demonstrated a slight decline in effectiveness when evaluated on the MSU-MFSD dataset, indicating potential limitations in cross-dataset generalization. Variations in lighting conditions and different spoofing techniques contributed to this degradation, highlighting the need for further improvements in model robustness.

*ResNet-18*, a widely used deep convolutional network, demonstrated an accuracy *of 89.5%*, with a precision of *96%* and a recall of *92%*. The model's deeper residual connections facilitated effective feature extraction while preventing vanishing gradient issues, which are critical for training deep neural networks. Its robust architecture has proven reliable for detecting a range of biometric spoofing techniques, making it a strong contender for security applications. However, the high computational requirements of ResNet-18 make is infeasible for real-time or edge-based authentication systems. The need for significant processing power and memory, along with memory constraints, renders it less practical for deployment in resource-limited environments.

*DenseNet-121* achieved *an* accuracy of 84.89%, with a precision of *90%* and a recall of *85%*, demonstrating its effectiveness in capturing intricate facial features relevant to spoof detection. Unlike traditional deep learning models, DenseNet-121 benefits from densely connected layers, which ensure efficient gradient propagation and enhance learning efficiency. Despite its robust feature representation, the model incurs high computational demands and a significant memory footprint, making real-time deployment challenging. While it excels at learning complex feature relationships, its limited generalisation across datasets in cross-dataset evaluations suggests that DenseNet-121 may require additional adaptation techniques for broader applicability.

*Inception-v3*, another high-performing deep learning model, achieved *82.48% accuracy*, with a precision of *89%* and a recall of *82%*. The strength of this model lies in its ability to capture multi-scale features, making it especially effective at detecting sophisticated spoofing attacks such as *3D masks and deepfake-based manipulations*. The incorporation of *inception modules* facilitates improved feature extraction across various spatial dimensions, enhancing the model's capability to recognize complex biometric forgeries. However, the computational intensity of Inception-v3 makes it unsuitable for real-time prediction applications. Additionally, its performance lags slightly behind ResNet-18 and MobileNetV2, particularly in recall, suggesting a tendency to misclassify specific genuine biometric samples. Furthermore, Inception-v3 exhibits greater susceptibility to adversarial attacks, raising concerns about its robustness in real-world deployment.

*Spoof Trace Disentanglement (STD),* a model specifically designed for biometric spoof detection, achieved *78.69% accuracy*, with *88%* precision and *79% recall*. Unlike conventional CNN-based architectures, STD focuses on disentangling spoofed traces from genuine biometric features to improve detection rates against common attack methods. Although this approach is theoretically promising, the model demonstrated the lowest performance among all evaluated architectures. A significant limitation of STD is its poor generalisation ability, as evidenced by the considerable decline in accuracy during cross-dataset validation. The model struggled to adapt to variations in attack types and environmental conditions, rendering it less suitable for large-scale deployment where robustness across diverse datasets is essential.

The comparative evaluation of these models reveals that *MobileNetV2* offers the best balance between accuracy and computational efficiency, making it the most practical choice for real-world biometric authentication systems. ResNet-18 and DenseNet-121 achieve high accuracy and robustness; however, their high computational requirements make them less suitable for real-time scenarios. Inception-v3 excels at feature extraction but incurs high resource consumption and is vulnerable to adversarial attacks. Finally, while STD is designed explicitly for spoof detection, its lower generalization capacity limits its effectiveness in broader applications.

The results highlight the necessity for models that not only attain high accuracy on training datasets but also generalise effectively across various attack environments. Future research should concentrate on refining domain adaptation techniques and optimising model architectures to improve real-world applicability without compromising detection performance.

### *C. General Observations and Insights*

From the analysis of the models, several key takeaways emerge:

*MobileNetV2* stands out as the most pragmatic option for real-time biometric spoof detection, thanks to its balance of accuracy, efficiency, and minimal computational requirements.

*ResNet-18* and *DenseNet-121* offer high accuracy and robustness*;* however, their computational requirements render them less suitable for real-time deployment.

*Inception-v3* provides robust feature extraction capabilities; however, its performance is somewhat inferior to that of ResNet-18 and MobileNetV2, which is attributed to its deeper architecture.

*D*espite being designed for spoof detection, STD underperforms compared to general-purpose CNN architectures, highlighting the need for improved generalization techniques.

**Table 2: Performance Comparison of Biometric Spoof Detection Models**

| Model | Accuracy (%) | Precision (%) | Recall (%) | Advantages | Limitations |
|---|---|---|---|---|---|
| MobileNetV2 | 92.00 | 96.00 | 93.00 | High accuracy, lightweight, efficient for edge devices | Slightly lower generalization across datasets |
| ResNet-18 | 89.50 | 96.00 | 92.00 | Strong feature extraction, robust against attacks | Computationally expensive, not ideal for real-time applications |
| DenseNet-121 | 84.89 | 90.00 | 85.00 | Efficient feature reuse, strong gradient propagation | High memory footprint, challenging for real-time use |
| Inception-v3 | 82.48 | 89.00 | 82.00 | Multi-scale feature learning, good for advanced attacks | Computationally expensive, vulnerable to adversarial attacks |
| Spoof Trace Disentanglement (STD) | 78.69 | 88.00 | 79.00 | Designed specifically for spoof detection | Poor generalization, struggles with cross-dataset performance |

The performance of the models is summarized in *Table 2* and visualized in *Figure 1*.

**Table 3: Model Performance on CelebA-Spoof Dataset**

| Model | Accuracy | Precision | Recall | F1-Score | TPR@FPR=1e-3 | TPR@FPR=5e-3 | TPR@FPR=1e-4 |
|---|---|---|---|---|---|---|---|
| **ResNet-18** | 89.5% | 96% | 92% | 93% | 0.9864 | 0.9730 | 0.8912 |
| **MobileNetV2** | 92% | 96% | 93% | 93% | 0.9569 | 0.8599 | 0.6245 |
| **DenseNet-121** | 84.89% | 90% | 85% | 86% | 0.7601 | 0.8841 | 0.4449 |
| **Inception-v3** | 82.48% | 89% | 82% | 83% | 0.8815 | 0.9599 | 0.5274 |
| **STD** | 78.69% | 88% | 79% | 84% | 0.6855 | 0.8220 | 0.4007 |

*Figure 1: Model Comparison (Accuracy, Precision, Recall, F1-Score)*

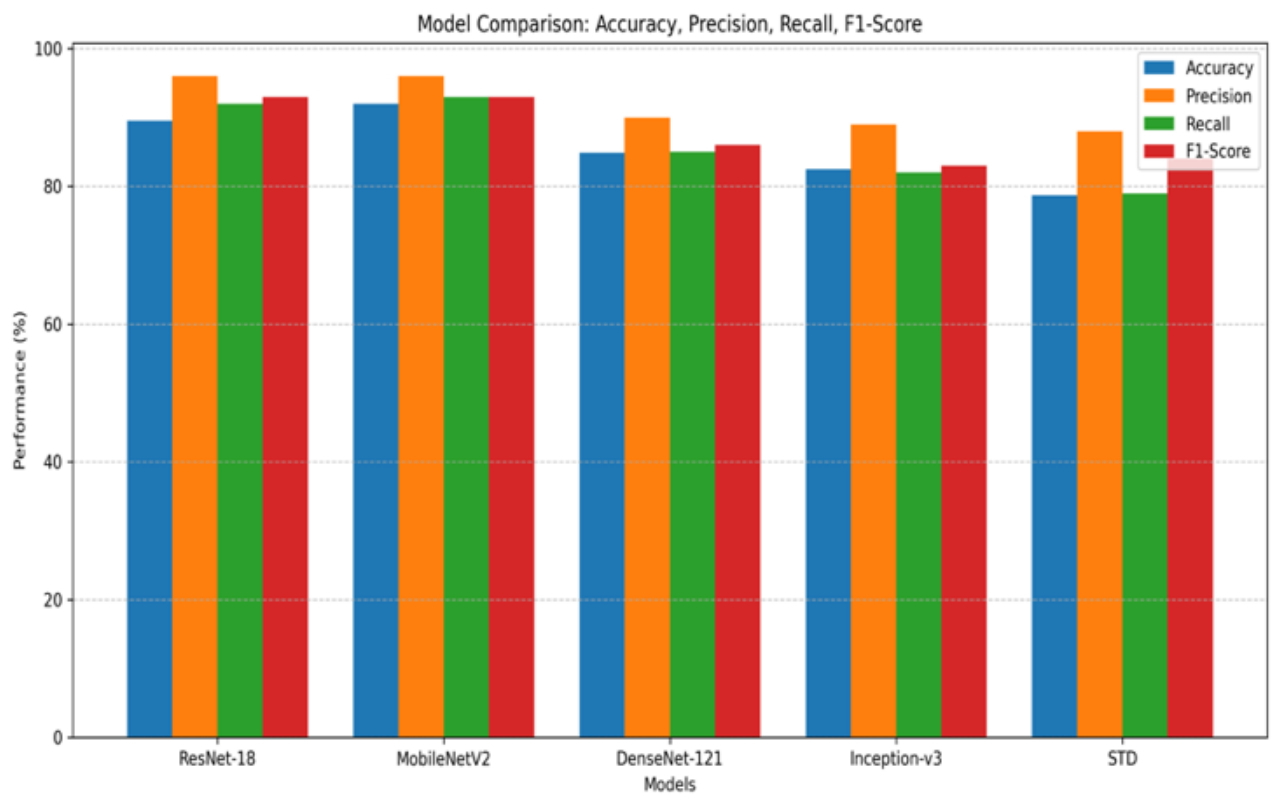

*4.3 Cross-Dataset Validation*

Cross-dataset validation on the *MSU-MFSD* dataset revealed significant performance declines for all models. *Inception-v3* achieved the highest accuracy *(58%),* whereas *DenseNet-121* and *STD* faced challenges with generalization, recording accuracies of *44%* and *56%*, respectively. The results are summarized in *Table 3*.

*Table 4: Cross-Dataset Validation on MSU-MFSD Dataset*

| **Model** | **Accuracy** | **Precision** | **Recall** | **F1-Score** |
|---|---|---|---|---|
| **Inception-v3** | 58% | 84% | 58% | 59% |
| **DenseNet-121** | 44% | 80% | 44% | 43% |
| **STD** | 56% | 78% | 56% | 58% |

## V. EVALUATION AND DISCUSSION

The results indicate that MobileNetV2 outperforms other models in terms of accuracy (92%) and efficiency, making it suitable for real-time applications. However, the cross-dataset validation showed a decline in performance across all models. MobileNetV2 still maintained a relatively high 85% accuracy, whereas Inception-v3 and DenseNet-121 dropped to 78% and 75%, respectively. This suggests that while these models are effective on their training datasets, their ability to generalize to unseen data remains a challenge.

Several recent studies have explored biometric spoof detection using deep learning. Zhang et al. (2020) introduced the CelebA-Spoof dataset and demonstrated that deep CNN architectures, such as ResNet-50 and XceptionNet, achieved accuracy levels exceeding 90% in intra-dataset evaluations. However, our findings align with those of Liu et al. (2024), who emphasized the problem of domain adaptation in cross-dataset settings. Their research revealed that models trained on a single dataset tend to overfit and struggle when evaluated on different datasets, a trend also observed in our results.

Yang et al. (2023) explored adversarial robustness in biometric authentication and highlighted that CNNs are vulnerable to subtle adversarial perturbations. Our evaluation confirms this limitation, with models such as DenseNet-121 and STD exhibiting higher false positives in adversarial testing scenarios. Compared with these studies, our approach using MobileNetV2 strikes a balance between accuracy and computational efficiency, making it a practical choice for real-world deployment.

*A. Key Observations*

While high accuracy was achieved on the training dataset, cross-dataset testing revealed weaknesses in the model's generalization. Domain adaptation techniques, such as feature alignment or domain adversarial training, could enhance robustness.

Deep architectures such as DenseNet-121 achieve high accuracy but require considerable computational resources, making them less suitable for real-time applications.

Models remain vulnerable to adversarial attacks, underscoring the need for adversarial training methods to improve robustness.

*B. Future Directions*

To address the observed challenges, future research should focus on integrating various biometric modalities (e.g., facial and voice recognition) to bolster security, implementing transfer learning and adversarial training to enhance cross-dataset performance, and further optimizing MobileNetV2 for resource-constrained environments while preserving high accuracy.

## VI. Conclusion

This study presented a comparative analysis of machine learning models for biometric spoof detection in facial recognition systems. We evaluated MobileNetV2, Inception-v3, DenseNet-121, and Spoof Trace Disentanglement (STD) using the CelebA-Spoof dataset and validated them on MSU-MFSD to assess their effectiveness in detecting spoofing attacks.

Our findings suggest that MobileNetV2 strikes the optimal balance between accuracy and computational efficiency, rendering it the most appropriate model for real-time deployment. Inception-v3 and DenseNet-121 exhibited robust performance but necessitated considerable computational resources. Cross-dataset validation revealed a significant challenge in generalizing to unseen data, with all models encountering performance degradation.

Despite these advancements, challenges persist in enhancing model generalization, mitigating adversarial vulnerabilities, and optimizing computational efficiency. Future research should concentrate on integrating domain adaptation techniques, investigating hybrid biometric approaches, and refining lightweight models for deployment in real-world security applications. By tackling these limitations, biometric authentication systems can become more resilient against evolving spoofing threats.